\documentclass[10pt,twocolumn,letterpaper]{article}

\usepackage{iccv}
\usepackage{times}
\usepackage{epsfig}
\usepackage{graphicx}
\usepackage{amsmath}
\usepackage{amssymb}
\usepackage{multirow}
\usepackage{booktabs}
\usepackage{authblk}

\usepackage[breaklinks=true,letterpaper=true,bookmarks=false]{hyperref}

\iccvfinalcopy  

\def\iccvPaperID{2601} 

\ificcvfinal\fi
\begin{document}

\title{Deep Mesh Reconstruction from Single RGB Images \\ via Topology Modification Networks}

\author[1]{Junyi Pan}
\author[2]{Xiaoguang Han}
\author[3]{Weikai Chen}
\author[1]{Jiapeng Tang}
\author[1]{Kui Jia\thanks{Corresponding author}}
\affil[1]{School of Electronic and Information Engineering, South China University of Technology}
\affil[2]{Shenzhen Research Institute of Big Data, the Chinese University of Hong Kong (Shenzhen)}
\affil[3]{USC Institute for Creative Technologies}

\maketitle
\thispagestyle{empty}
\pagestyle{empty}

\begin{abstract}
\vspace{-0.2cm}
Reconstructing the 3D mesh of a general object from a single image is now possible thanks to the latest advances of deep learning technologies. 
However, due to the nontrivial difficulty of generating a feasible mesh structure, the state-of-the-art approaches \cite{kato2018neural,wang2018pixel2mesh} often simplify the problem by learning the displacements of a template mesh that deforms it to the target surface. Though reconstructing a 3D shape with complex topology can be achieved by deforming multiple mesh patches, it remains difficult to stitch the results to ensure a high meshing quality. In this paper, we present an end-to-end single-view mesh reconstruction framework that is able to generate high-quality meshes with complex topologies from a single genus-0 template mesh. 
The key to our approach is a novel progressive shaping framework that alternates between mesh deformation and topology modification. While a deformation network predicts the per-vertex translations that reduce the gap between the reconstructed mesh and the ground truth, a novel topology modification network is employed to prune the error-prone faces, enabling the evolution of topology. By iterating over the two procedures, one can progressively modify the mesh topology while achieving higher reconstruction accuracy. 
Moreover, a boundary refinement network is designed to refine the  boundary conditions to further improve the visual quality of the reconstructed mesh.
Extensive experiments demonstrate that our approach outperforms the current state-of-the-art methods both qualitatively and quantitatively, especially for the shapes with complex topologies.
\end{abstract}


\section{Introduction}

Image-based 3D reconstruction plays a fundamental role in a variety of tasks in computer vision and computer graphics, such as robot perception, autonomous driving, virtual/augmented reality,  \textit{etc.}
Conventional approaches mainly leverage the stereo correspondence based on multi-view geometry but are restricted to the coverage provided by the input views. 
Such requirement renders single-view reconstruction particularly difficult due to the lack of correspondence and large occlusions.
With the availability of large-scale 3D shape dataset~\cite{chang2015shapenet}, shape priors can be efficiently encoded in a deep neural network, enabling faithful 3D reconstruction even from a single image.
While a variety of 3D representations, e.g. voxels~\cite{choy20163d,tulsiani2017multi,yan2016perspective} and point cloud~\cite{fan2017point,yang2018foldingnet}, have been explored for single-view reconstruction, triangular mesh receives the most attentions as it is more desirable for a wide range of real applications and capable of modeling geometric details. 

\begin{figure}
    \centering
    \includegraphics[scale=0.45]{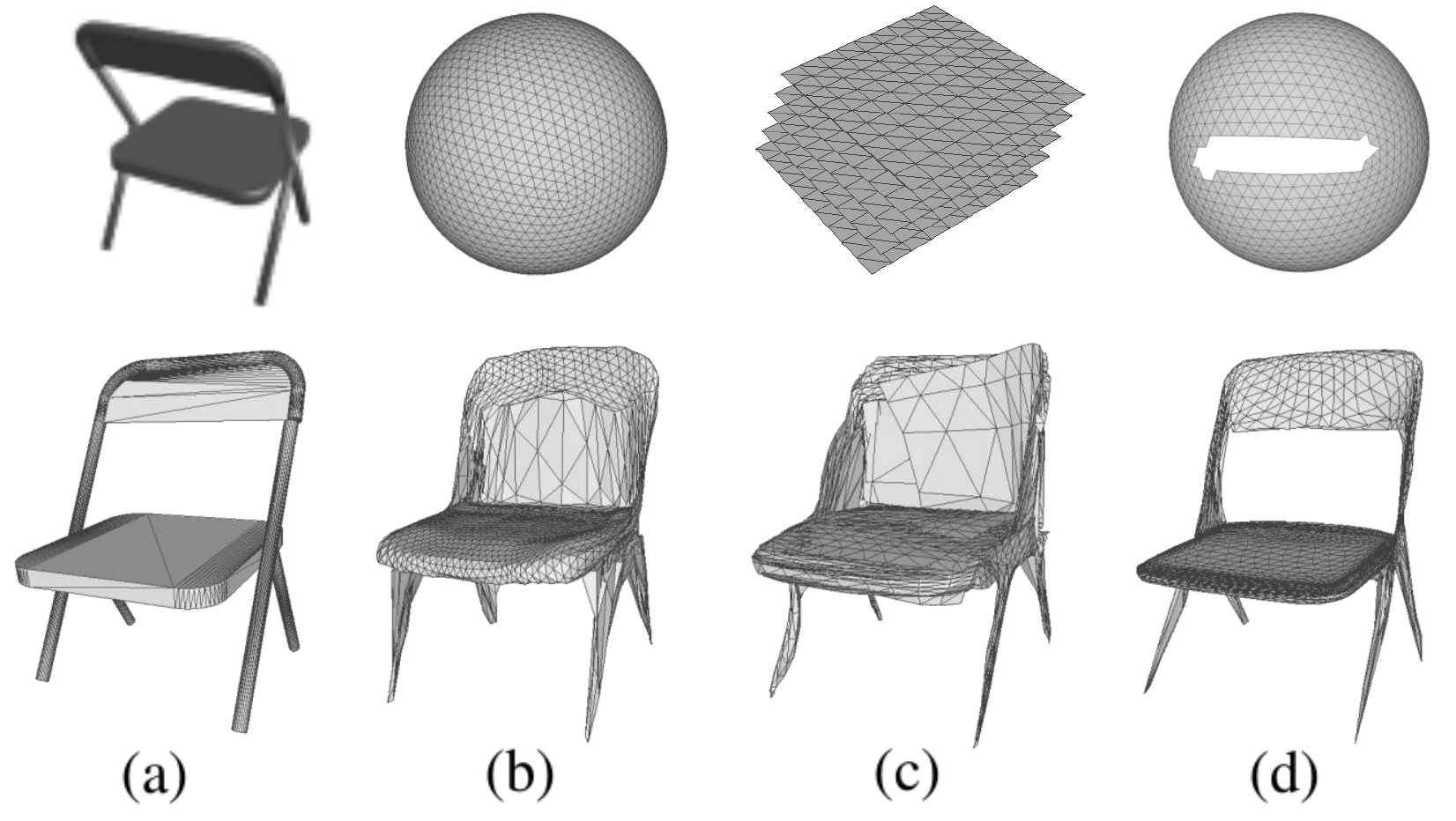}
    \vspace{-0.1cm}
    \caption{
    Given a single image of an object (a) as input, the existing mesh-deformation based learning approaches \cite{groueix2018atlasnet} can not well capture the complex topology, regardless of a single (b) or multiple template meshes (c). In contrast, our proposed method is capable of updating the topologies dynamically by removing faces in the initial sphere mesh and achieves better reconstruction results (d).}
    \label{fig:show}
\end{figure}


Recent progresses in single-view mesh reconstruction~\cite{wang2018pixel2mesh,groueix2018atlasnet} propose to reconstruct a 3D mesh by deforming a template model based on the perceptual features extracted from the input image. 
Though promising results have been achieved, the reconstructed results are limited to the identical topological structure with the template model, leading to large reconstruction errors when the target object has a different topology (cf. Figure~\ref{fig:show} (b)). 
Although it is possible to approximate a complex shape with non-disk topology by deforming multiple patches to cover the target surface, there remain several drawbacks that limit its practical usability.
Firstly, the reconstructed result is composed of multiple disconnected surface patches, leading to severe self-intersections and overlaps that require tedious efforts to remove the artifacts. 
Secondly, as obtaining a high-quality global surface parameterization remains a challenging problem, it is nontrivial to generate a proper atlas that can cover the surface with low distortion, only based on a single image.
Lastly, it is difficult to determine an appropriate number of surface patches that adapts to varying shapes. 
In this work, we strive to generate the 3D mesh with complex topology from a single genus-0 template mesh. Our key idea is a mechanism that dynamically modifies the topology of the template mesh by face pruning, targeting at a trade-off between the deformation flexibility and the output meshing quality. The basic model for deformation learning is a cascaded version of AtlasNet ~\cite{groueix2018atlasnet} that predicts per-vertex offsets instead of positional coordinates. Starting from an initial mesh $M_0$, we first apply such deformation network and obtain a coarse output $M_1$. Then, the key problem is to determine which faces on $M_1$ to remove. To this end, we propose to train an error-prediction network that estimates the reconstruction error (i.e. distance to the ground truth) of the reconstructed faces on $M_1$. The faces with large error would be removed to achieve better reconstruction accuracy. However, it remains nontrivial to determine a proper pruning threshold and to guarantee the smoothness of the open boundaries introduced by the face pruning.
Towards this end, we propose two strategies to address these issues: 1) a progressive learning framework that alternates between a mesh deformation network, which reduces the reconstruction error, and a topology modification network that prunes the faces with large approximation error; 
2) a boundary refinement network that 
imposes smoothness constraints on the boundary curves, to  
refine the boundary conditions. 
Both qualitative and quantitative evaluations demonstrate the superiority of our approach over the existing methods, in terms of both the reconstruction accuracy and the meshing quality. As seen in Figure~\ref{fig:show}, the proposed method is able to better capture the complex topology with a single sphere template mesh while achieving better meshing quality compared to the state-of-the-art AtlasNet~\cite{groueix2018atlasnet}.

In summary, our main contributions are:
 \begin{itemize}
   \item The first end-to-end learning framework for single-view object reconstruction that is capable of modeling complex mesh topology from a single genus-0 template mesh.
   \item A novel topology modification network, which can be integrated into other mesh learning frameworks.  
   \item We demonstrate the advantage of our approach over the state-of-the-arts in terms of both reconstruction accuracy and the meshing quality.  
\end{itemize}

\section{Related Works}

Reconstructing 3D surfaces from color images has been investigated since the very beginning of the field \cite{roberts1963machine}. 
To infer 3D structures from 2D images, conventional approaches mainly leverage the stereo correspondences from multi-view geometry \cite{hartley2003multiple,furukawa2010accurate}.
Though high-quality reconstruction can be achieved, stereo based approaches are restricted to the coverage provided by the multiple views and specific appearance models that cannot be generalized to non-lambertian object reconstruction.
Hence, learning-based approaches have stood out as the major trend in recent years thanks to its scalability to single or few images.

With the success of deep neural network and the availability of large-scale 3D shape collections, e.g. ShapeNet~\cite{chang2015shapenet}, deep learning-based 3D shape generation has made great progress. In order to replicate the success of 2D convolutional neural network to 3D domain, various forms of 3D representations have been explored. As a natural extension of 2D pixels, volumetric representation has been widely used in recent works on 3D reconstruction \cite{tatarchenko2017octree,tulsiani2017multi,yan2016perspective,choy20163d,huang2018deep,varol2018bodynet,zhou2018hairnet,han2017high,yamaguchi2018high} due to its simplicity of implementation and compatibility with convolutional neural network.
However, deep voxel generators are constrained by its resolution due to the data sparsity and computation cost of 3D convolution.
As a flexible form of representing a 3D structure, point cloud has become another major alternative for 3D learning \cite{qi2017pointnet,chen2018deep} and shape generation \cite{fan2017point,lin2018learning,achlioptas2017learning,insafutdinov2018unsupervised,yang2018foldingnet,mandikal20183dlmnet} due to its high memory efficiency and simple and unified structure.
Though enjoying the flexibility to match 3D shape with arbitrary topology, point cloud is not a well suited for imposing geometry constraints, which are critical for ensuring smoothness and appealing visual appearance of the reconstructed surface. 
Implicit field based 3D reconstruction approaches~\cite{mescheder2018occupancy,michalkiewicz2019deep,chen2018learning,natsume2019siclope,huang2018deep} share the similar advantages with point cloud representation in providing good trade-offs across fidelity, flexibility and compression capabilities. Yet it also remains difficult to regularize the generation of a volumetric implicit field to achieve specific geometry properties.

\begin{figure*}
    \centering
    \includegraphics[width=0.95\linewidth]{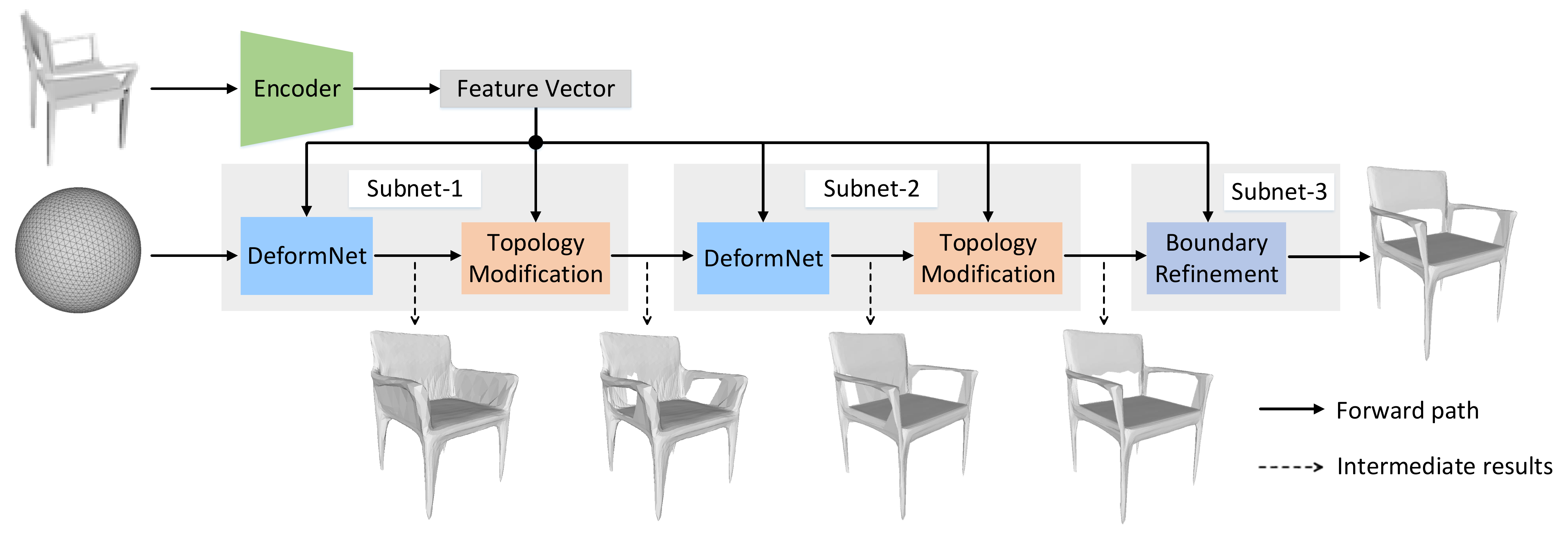}
    \vspace{-0.2cm}
    \caption{The overview of our pipeline. Given an input image, we first employ multiple mesh deformation
    and topology modification modules to progressively deform the mesh vertices and update the topologies
    to approximate the target object surface. 
    A module of boundary refinement is then adopted to refine the
    boundary conditions.
    }
    
    \label{fig:pip}
\end{figure*}

In contrast, mesh representation is more desirable for real applications since it can model fine shape details and is compatible with various geometry regularizers.
Due to the complexity of modifying the mesh topology, most mesh learning approaches strive to obtain a target shape by deforming a template mesh \cite{groueix2018atlasnet,wang2018pixel2mesh,pontes2017image2mesh,kanazawa2018learning,pan2018residual} via the learned shape prior.
More recently, the advances in differentiable renderer \cite{liu2019softras,kato2018neural} have proposed to train a mesh generator based on rendering loss, eliminating the need of 3D supervision.
However, no prior approaches can dynamically modify the topology of the template mesh, while we propose the first topology modification network that is able to generate meshes with complex topologies from a  genus-0 3D model.


\section{Topology-adaptive Mesh Reconstruction}
\paragraph{Overview.}
Given a single image $I$ of an object, we attempt to reconstruct the surface $S$ of it. We adopt the triangular mesh as a natural and flexible discretization of the target surface. A mesh is typically defined by $M=(V,E,T)$, where $V\in\mathbb{R}^{3}$ is the set of mesh vertices, $E$ is the set of edges connecting the neighboring vertices, and $T$ is the set of triangles enclosed by the connected edges. 
To reconstruct the triangular mesh representation of an object, one could choose to deform a template mesh to approximate the target surface. Nevertheless, the existing deformation-based mesh reconstruction approaches, such as ~\cite{wang2018pixel2mesh,groueix2018atlasnet,kato2018neural}, are not allowed to update the faces-to-vertices relationships and thus are restricted by the predefined topology. In order to overcome this limitation, we propose an end-to-end learning pipeline, consisting of three modules, to progressively modify the coordinates and connectivity of the vertices on a predefined mesh $M_0$. 
To be specific, the mesh deformation module is adopted to map the vertices on $M_0$ to the target surface while maintain the connectivity over them; the topology modification module is developed to update the connection relationship between the vertices by pruning the faces which deviate from the ground truth; the boundary refinement module is designed to refine the open boundaries introduced by face pruning. Note that the mesh deformation and topology modification are performed in an alternative manner to gradually recover the overall shape and topology of the target object.


\paragraph{Network structure.} We propose a progressive structure to deform a template mesh $M_0$ to fit the target surface $S$. In our implementation, $M_0$ is instantiated as a sphere mesh with $2562$ vertices. 
Figure \ref{fig:pip} illustrates the overall pipeline. We leverage an encoder-decoder network for shape generation. On the encoder side, the input image is fed into ResNet-18 \cite{he2016deep} to extract a $1024$-dimensional feature vector $x$.
The decoder contains three successive subnets. 
Each of the first two subnets consists of a mesh deformation module and a topology modification module, and the last subnet comprises a single boundary refinement module.
Note that each mesh deformation module predicts the per-vertex offset, which can be added to the input mesh to obtain the reconstructed result.
The topology modification module then estimates the reconstruction error of the outcome of the preceding deformation module and removes the faces with large error in order to update the mesh topology.
Finally, the boundary refinement module enhances the smoothness of the open boundaries to further improve the
visual quality.

\subsection{Mesh DeformNet}
Our mesh deformation module consists of a single multi-layer perceptron (MLP). Specifically, the MLP is composed of four fully-connected layers of size $1024$, $512$, $256$, $128$ with non-linear activation ReLU on the first three layers and \textit{tanh} on the final output layer. 
Given an initial mesh $M$ and the shape feature vector $x$ that contains the prior knowledge of the object, we replicate the vector $x$ and concatenate it with the matrix containing all the vertices of $M$ before feeding them into the MLP. The MLP performs the affine transformation on each vertex of $M$ and generates the vertex displacements.
Note that we choose to predict the offsets instead of directly regressing the coordinates. Such a design paradigm enables more accurate learning of fine geometric details with even less training time.

\subsection{Topology Modification}
\begin{figure}[t]
    \centering
    \includegraphics[scale=0.25]{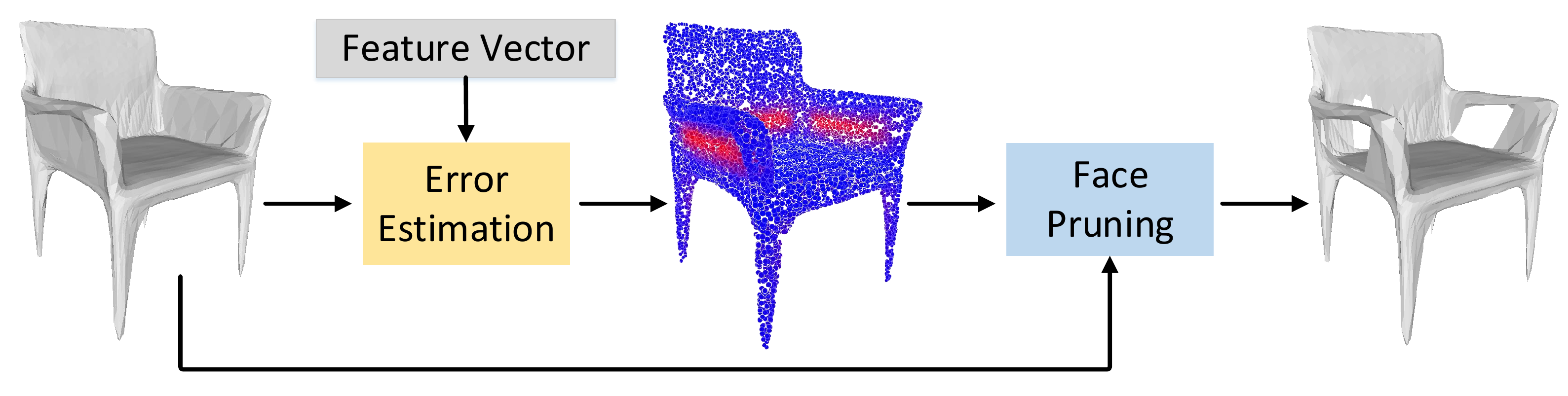}
    \vspace{-0.1cm}
    \caption{Topology Modification Network. The red color indicates the sampled points with higher estimated errors.}
    \label{fig:tmnet}
\end{figure}

To generate objects with various topologies, it is necessary to modify the faces-to-vertices relationship dynamically. Towards this goal, we propose a topology modification network that updates the topological structure of the reconstructed mesh by pruning the faces which deviate significantly from the ground truth.
The topology modification network is illustrated in Figure \ref{fig:tmnet},
which includes two components: error estimation and face pruning.

\subsubsection{Error Estimation}
\label{sec:errorEst}

To perform face pruning, it is key to locate the triangle faces that have large reconstruction errors at test time.
We propose an error estimation network that predicts the per-face errors of the reconstructed mesh from the preceding mesh deformation network.
It leverages a similar architecture with that of the mesh deformation network.  
In particular, we sample points randomly on the faces of the predicted mesh $M$ and concatenate the replicated shape feature vector $x$ with the matrix containing all the sampling points. 
The MLP takes as input the feature matrix and predicts the per-point errors (distances to the ground truth). 
The final error for each triangle face is obtained by averaging the predicted errors for all the sampling points of the triangle face.


\subsubsection{Face Pruning}
\label{sec:pruning}

Given the estimated error for each triangle face, we then apply a thresholding strategy that removes the faces whose estimated errors are beyond the predefined threshold to update the mesh topology. 
However, to obtain a properly pruned mesh structure, the threshold $\tau$ needs to be carefully configured: a higher value of $\tau$ tends to generate reconstructions with higher errors while a low decision threshold may remove too many triangles and destroy the surface geometry of the generated mesh.
To this end, we propose a progressive face pruning strategy that removes error-prone faces in a coarse-to-fine fashion.
In particular, we set a higher value for $\tau$ at the first subnet and decrease it by a constant factor at the subsequent subnet.
Such a strategy enables the face pruning to be performed in a much more accurate manner.

\subsection{Boundary Refinement}
\label{sec:boundary}

As shown in Figure~\ref{fig:pip}, a naive pruning of triangles will introduce jagged boundaries that adversely impact the visual appearance. 
To prevent such artifacts and further improve the visual quality of the reconstructed mesh, we design a boundary refinement module to enhance the smoothness of the open boundaries. 
It is similar to the mesh deformation module but only predicts the displacement with respect to each input boundary vertex. Note that each boundary vertex is only allowed to move on the 2D plane established by the two boundary edges that intersect at the vertex.
We further propose a novel regularization term which penalizes the zigzags by enforcing the boundary curves to stay smooth and consistent. The boundary regularizer is defined as follows:
\begin{eqnarray}
\begin{aligned}
    {\cal{L}}_{bound} = \sum_{x \in E} \Bigg \| \sum_{p \in \mathcal{N}(x)}\frac{(x-p)}{\|x-p\|_{}}\Bigg \|,
\end{aligned}
\end{eqnarray}
where $\{x \in E\}$ is the set of vertices which lie on the open boundary and $\{p \in \mathcal{N}(x)\}$ is the set of neighboring vertices of $x$ on the boundary.


\subsection{Training Objectives}


Our network is supervised by a hybrid of losses.
For mesh deformation and boundary refinement, we employ the commonly-used Chamfer distance (CD) for measuring the discrepancy between the reconstructed result and the ground truth. 
The error estimation network is trained by the quadratic loss for regressing the reconstruction errors. The boundary regularizer is proposed to guarantee the smoothness of the boundary curves. Besides,
we also apply a combination of geometry constraints to regularize the smoothness of the mesh surface during mesh deformation. 

\paragraph{CD loss.}
The CD measures the nearest neighbor distance between two point sets. In our setting, we minimize the two directional distances between the point set randomly sampled from the generated mesh $M$ and the ground truth point set. The CD loss is defined as:
\begin{eqnarray}\label{cd}
\begin{aligned}
{\cal{L}}_{cd} = \sum_{x\in M} \min_{y\in S} \|x-y\|_2^2 + \sum_{y\in S}\min_{x\in M}\|x-y\|_2^2,
\end{aligned}
\end{eqnarray}
where $\{x \in M\}$ and $\{y \in S\}$ are respectively the point sets sampled from the generated mesh $M$ and the ground truth surface $S$. For each point, CD finds the nearest point in another point set, and sums the squared distances up.

\begin{figure*}[t]
    \centering
    \includegraphics[width=0.93\linewidth]{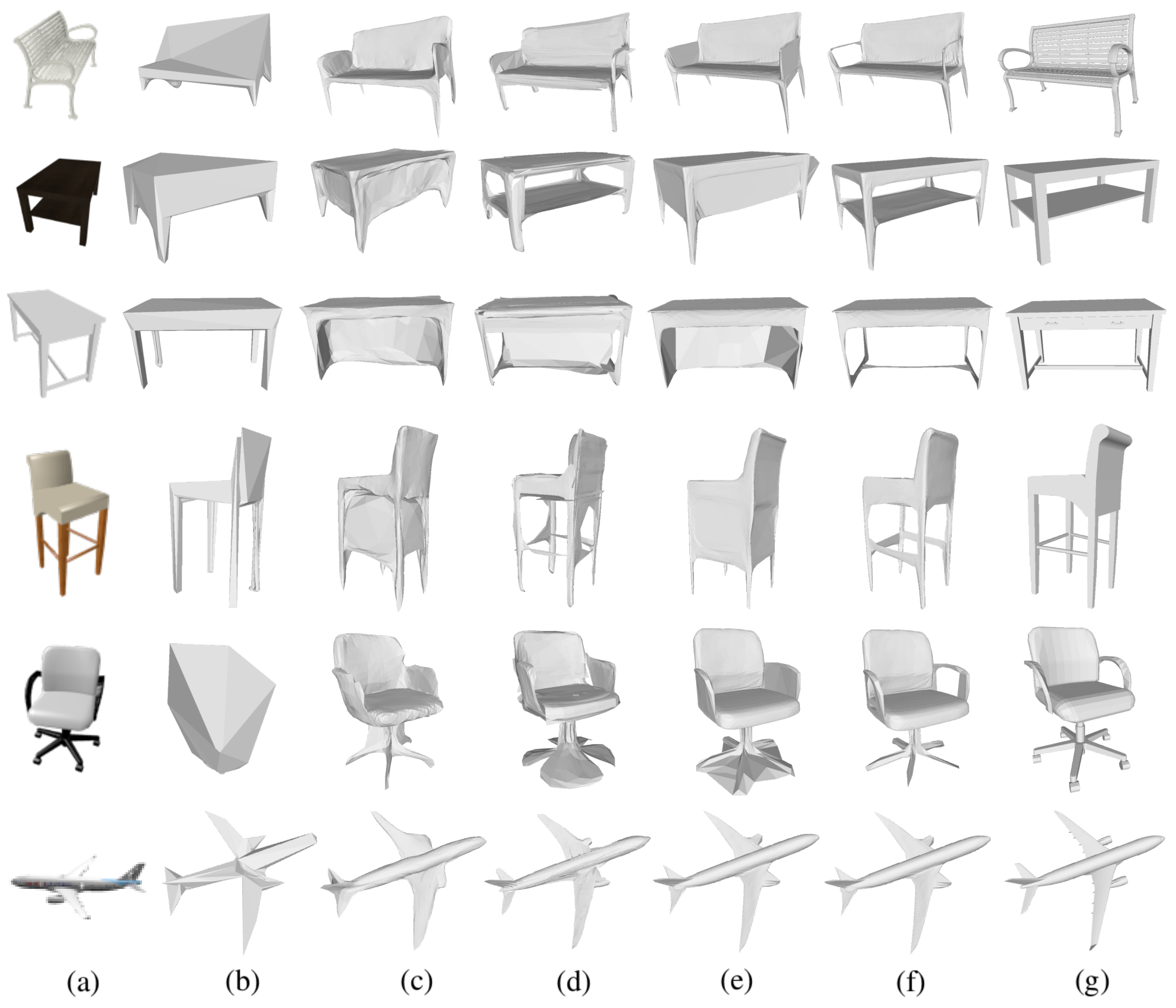}
    \vspace{-0.1cm}
    \caption{Qualitative results. (a) Input image; (b) N3MR; (c) Pixel2Mesh; (d) AtlasNet-25; (e) Baseline; (f) Ours;
    (g) Ground truth.}
    \label{fig:com}
\end{figure*}

\paragraph{Error estimation loss.}
We adopt the quadratic loss to train our error estimation network, which is defined as:
\begin{eqnarray}
\begin{aligned}
{\cal{L}}_{error} =  \sum_{x \in M} |f_{e}(x) - {e}_{x} |^{2},
\end{aligned}
\end{eqnarray}
where  $\{x \in M\}$ is the point set sampled from the generated mesh $M$, $f_{e}$ is the error estimation network, and $e_{x}$ is
the corresponding ground truth error.

\paragraph{Geometry regularizers.}
For the mesh deformation module, since the CD loss does not take into account the connectivity of mesh vertices, the predicted mesh could suffer from severe flying vertices and self-intersections. 
To improve the smoothness of the mesh surface, we add several geometry regularizers. 
We employ three regularization techniques defined in \cite{wang2018pixel2mesh,kato2018neural}: the normal loss  ${\cal{L}}_{normal}$ which measures the normal consistency between the generated mesh and ground truth, the smoothness loss ${\cal{L}}_{smooth}$ which flattens the intersection angles of the triangle faces and supports the surface smoothness, and the edge loss ${\cal{L}}_{edge}$ which penalizes the flying vertices and overlong edges to guarantee the high quality of recovered 3D geometry.

The final training objective of our system is defined as: 

\begin{eqnarray}
\begin{aligned}
{\cal{L}} ={\cal{L}}_{cd} + \lambda_{1}{\cal{L}}_{error} +  \lambda_{2} {\cal{L}}_{bound}  +\lambda_{3}{\cal{L}}_{normal}  \\ +
\lambda_{4}{\cal{L}}_{smooth} +  \lambda_{5}{\cal{L}}_{edge},
\end{aligned}
\end{eqnarray}

\noindent where $\lambda_{i}$ are hyper-parameters weighting the importance of each loss term.

\section{Experiments}

\setlength{\tabcolsep}{4pt}
\renewcommand\arraystretch{1}
\begin{table*}[t]
    \centering
    \begin{tabular}{*{11}{c}}
    
         \toprule
         \multirow{2}{*}{Category} & \multicolumn{5}{c}{CD}  & \multicolumn{5}{c}{EMD} \\
         \cmidrule(lr){2-6} \cmidrule(lr){7-11}
         &N3MR & Pixel2Mesh & AtlasNet-25 & Baseline & Ours  &N3MR & Pixel2Mesh & AtlasNet-25 & Baseline & Ours\\
         \midrule
         plane  & 3.550 & 2.130 & 1.566 & 1.433 &\textbf{1.390} & 10.163 & 8.859 & 11.268 & 8.524 &\textbf{8.371}\\
         bench  & 10.865 & 3.107 & 2.239 & 2.950 &\textbf{2.172} & 14.101 & 10.075 &9.808 & 9.828 &\textbf{8.713}\\
         chair  & 15.891 & 4.787 & 3.796 & 4.325 &\textbf{3.064} & 17.246 & 13.498 & 11.956 & 13.313 &\textbf{10.383}\\
         table  & 13.438 & 5.339 & 4.647 & 4.798 &\textbf{3.616} & 15.697 & 11.452 & 11.562 & 11.837 &\textbf{9.604}\\
         firearm & 3.230  & 2.290 & 1.489 &1.145  &\textbf{1.142} & 13.581 &    8.590 & 8.711 & 8.297 &\textbf{8.226}\\
         \midrule
         mean   & 9.355 & 3.531 & 2.747 & 2.930 &\textbf{2.277} & 14.158 & 10.495 &10.661& 10.360 &\textbf{9.059} \\
         \bottomrule
         &  \\
         & 
    \end{tabular}
    \vspace{-0.6cm}
    \caption{Quantitative comparison with the state-of-the-art methods.
The CD and EMD are computed on
$10,000$ points sampled from the generated mesh after performing ICP alignment with the ground truth.
The CD is in units of $10^{-3}$ and the EMD is in units of $10^{-2}$. 
}
    \label{tab:cd_emd}
\end{table*}

\paragraph{Dataset.}

Our experiments are performed on the 3D models collected from five categories in
the ShapeNet \cite{chang2015shapenet} dataset. 
To ensure fair comparisons with the existing methods, we adopt the experiment setup in \cite{groueix2018atlasnet}. 
We use the rendered images provided by \cite{choy20163d} as the inputs, where each
3D model corresponds to $24$ RGB images. 
For each 3D shape, $10,000$ points are uniformly sampled on the surface as the ground truth.

\paragraph{Implementation details.}
The input images all have the same resolution of $224\times224$.
We first train each subnet
separately with fixing other components using a batch size of 16 with a learning rate of $1e-3$ (dropped to $1e-4$
after $200$ epochs) for $300$ epochs. The entire network is then fine-tuned in an end-to-end manner.
The values of hyper-parameters used in Equation (4) are $\lambda_1 = 1.0$, $\lambda_2 = 0.5$, $\lambda_3 = 1e-2$,
$\lambda_4 = 2e-7$, $\lambda_5 = 0.1$. The threshold $\tau$ for face pruning is set to be $0.1$ at the first subnet and decreased by a factor of $2$ at the subsequent subnet.

\subsection{Comparisons with the State-of-the-arts}

We first compare the performance of our approach with three state-of-the-art methods for single view 3D mesh reconstruction, including Neural 3D Mesh Renderer \cite{kato2018neural} (N3MR), Pixel2Mesh \cite{wang2018pixel2mesh} and AtlasNet \cite{groueix2018atlasnet} with 25 patches (AtlasNet-25). We also compare with the baseline approach which refers
to our framework without the topology modification and boundary refinement module.

\paragraph{Qualitative comparisons.}
The visual comparison results are shown in Figure~\ref{fig:com}. 
While N3MR can reconstruct the rough shapes, it fails to capture the fine details of the geometry and is not able to model surface with non-disk topology. 
Pixel2Mesh performs generally better than N3MR in terms of the capability of modeling the fine structures.
However, as Pixel2Mesh employs a similar mesh deformation strategy, it struggles to reconstruct shapes with complex topologies, especially for the chairs and tables.
Our baseline approach also has the same problem as the topology modification module is not applied.
Thanks to the use of multiple squares as the template model, AtlasNet-25 can generate meshes with various topologies. 
However, it suffers from severe self-intersections and overlapping and still fails to reconstruct
some instances with more complex topologies, e,g, the desk in the third row and the chair in the fifth row. 
In comparison, our approach has outperformed the other approaches in terms of visual quality.
We are able to generate meshes with complex topologies while maintaining high reconstruction accuracy thanks to the topology-modification modules. 
In addition, our method scales well to the shapes with simple topologies.
For the objects that can be well reconstructed from a template sphere (e.g. the plane),
the spherical topology is faithfully preserved.

\setlength{\tabcolsep}{10pt}
\begin{table}
    \centering
    \begin{tabular}{c|c|c}
        \hline
         Method & CD    & EMD  \\
         \hline
         AtlasNet-25 (PSR)  & 3.430 & 12.574 \\
         Ours (PSR)  & \textbf{2.304} & \textbf{10.415} \\
         \hline
    \end{tabular}
    \vspace{0.1cm}
    \caption{Quantitative comparison with AtlasNet-25 after PSR.  }
    \label{tab:psr}
\end{table}

\begin{figure}
    \centering
    \includegraphics[width=0.9\linewidth]{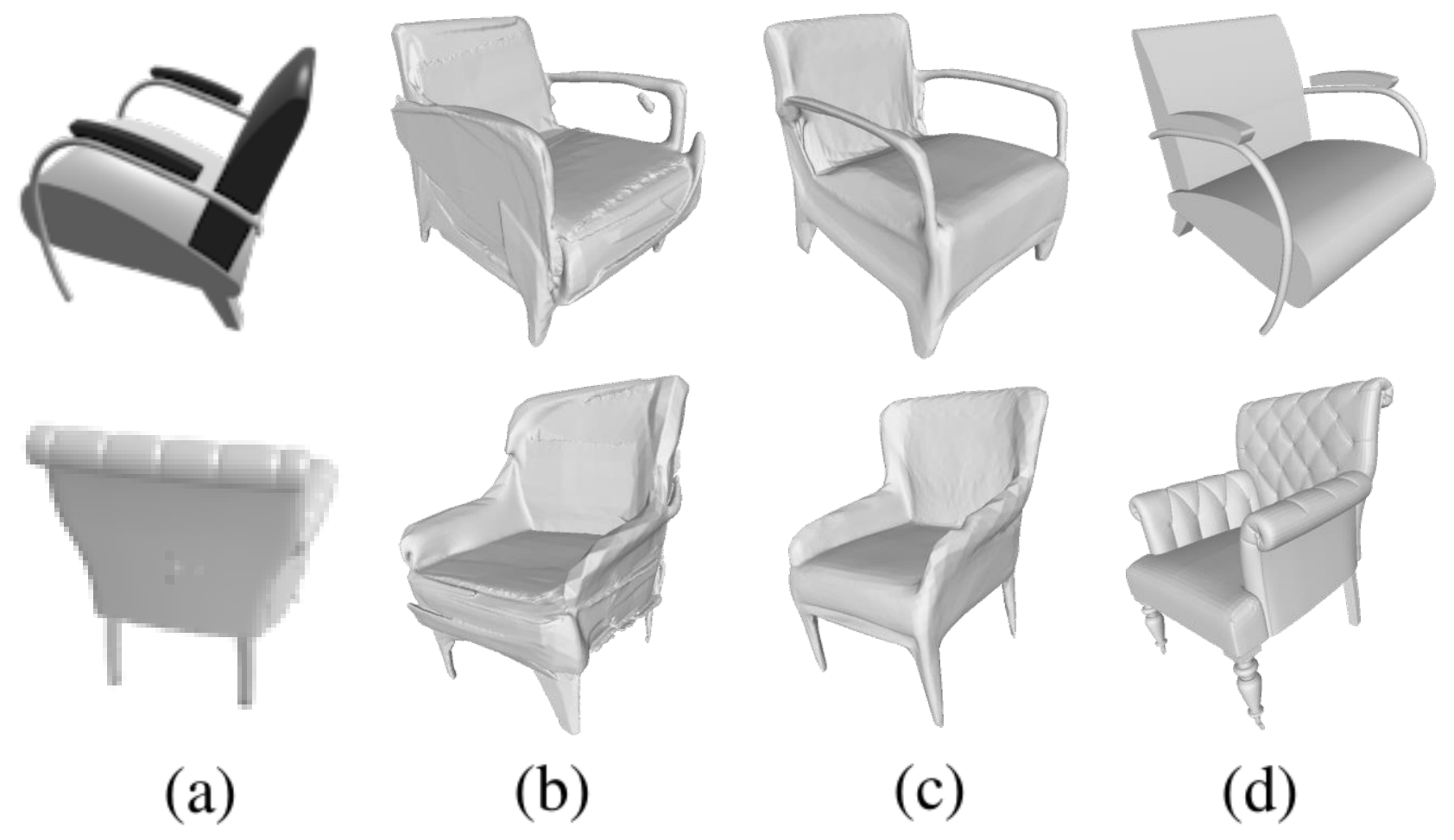}
    \caption{Qualitative comparison with AtlasNet-25 after PSR. (a): Input images; (b): AtlasNet-25;
    (c): Ours; (d): Ground truth.}
    \label{fig:psr}
\end{figure}

\paragraph{Quantitative comparisons.}

We adopt the widely used Chamfer Distance (CD) and Earth Mover's Distance (EMD)
to quantitatively evaluate the results. 
Both metrics are
computed between the ground truth point cloud and $10,000$ points uniformly sampled 
from the generated mesh. 
Since the outputs of Pixel2Mesh \cite{wang2018pixel2mesh} are non-canonical,
we align their predictions to the canonical ground truth by using the pose metadata available in the dataset. Additionally,
we apply the iterative closest point algorithm (ICP)~\cite{besl1992method} on all the results for finer alignment with the ground truth.
The quantitative comparison results are shown in 
Table \ref{tab:cd_emd}.  
Our approach consistently outperforms the state-of-the-art methods in both metrics over all five categories,
especially on the models with complex topologies (e.g. chair and table).

\begin{figure}
    \centering
    \includegraphics[width=0.92\linewidth]{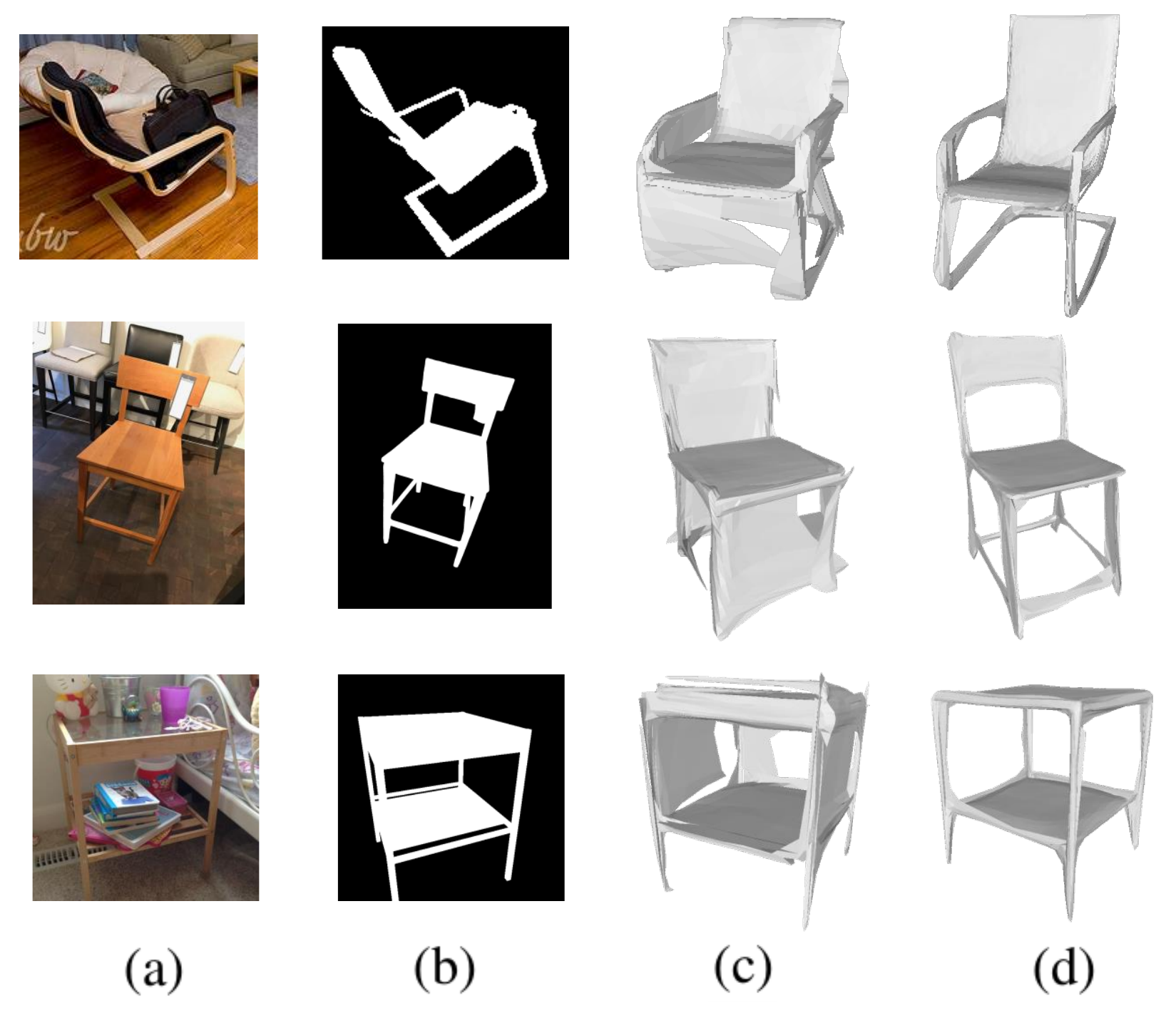}
    \caption{Qualitative comparison with AtlasNet-25 on real images. (a): Input images; (b): Input masks;
    (c): AtlasNet-25; (d): Ours.}
    \label{fig:pix3d}
\end{figure}

\paragraph{Poisson surface reconstruction.}

Although our method can generate visually appealing meshes with smooth surfaces and
complex topologies, it still has the inherent drawback of producing open surfaces due to the face pruning operations.
To avoid the above mentioned drawbacks, one could densely sample the surface
and reconstruct the mesh from the obtained point cloud.
In particular, we first sample $100,000$ points, together with their oriented normals, from the reconstructed surface
and then apply Poisson surface reconstruction (PSR)~\cite{kazhdan2013screened} to produce a closed triangle mesh.
To evaluate the performance of applying PSR on our results, we quantitatively compare with AtlasNet-25. Specifically, for both methods, we randomly selected $20$ shapes from
the chair category, and run the PSR algorithm to get the corresponding closed meshes.
In Table~\ref{tab:psr}, we show the quantitative comparisons measured in CD and EMD.
As seen from the results, our approach generates more accurate results under both measurements. 
We show the visual comparisons in Figure~\ref{fig:psr}.
Note that we generate meshes with significantly fewer artifacts and correct topologies 
compared to the AtlasNet, proving the better meshing quality of our method.

\paragraph{Reconstructing real-world objects.}

To qualitatively evaluate the  generalization performance of our method on the real images, we test
our network on the Pix3D \cite{sun2018pix3d} dataset by using the model trained on the ShapeNet \cite{chang2015shapenet}. Figure~\ref{fig:pix3d} shows the results 
reconstructed by our method and AtlasNet, where the objects in the images are manually segmented.
Our approach is still able to faithfully reconstruct a variety of objects with complex topologies and achieves better accuracy compared against AtlasNet, indicating that our method scales reasonably well on the real images.

\subsection{Ablation Studies}

\begin{figure}
    \centering
    \includegraphics[width=0.9\linewidth]{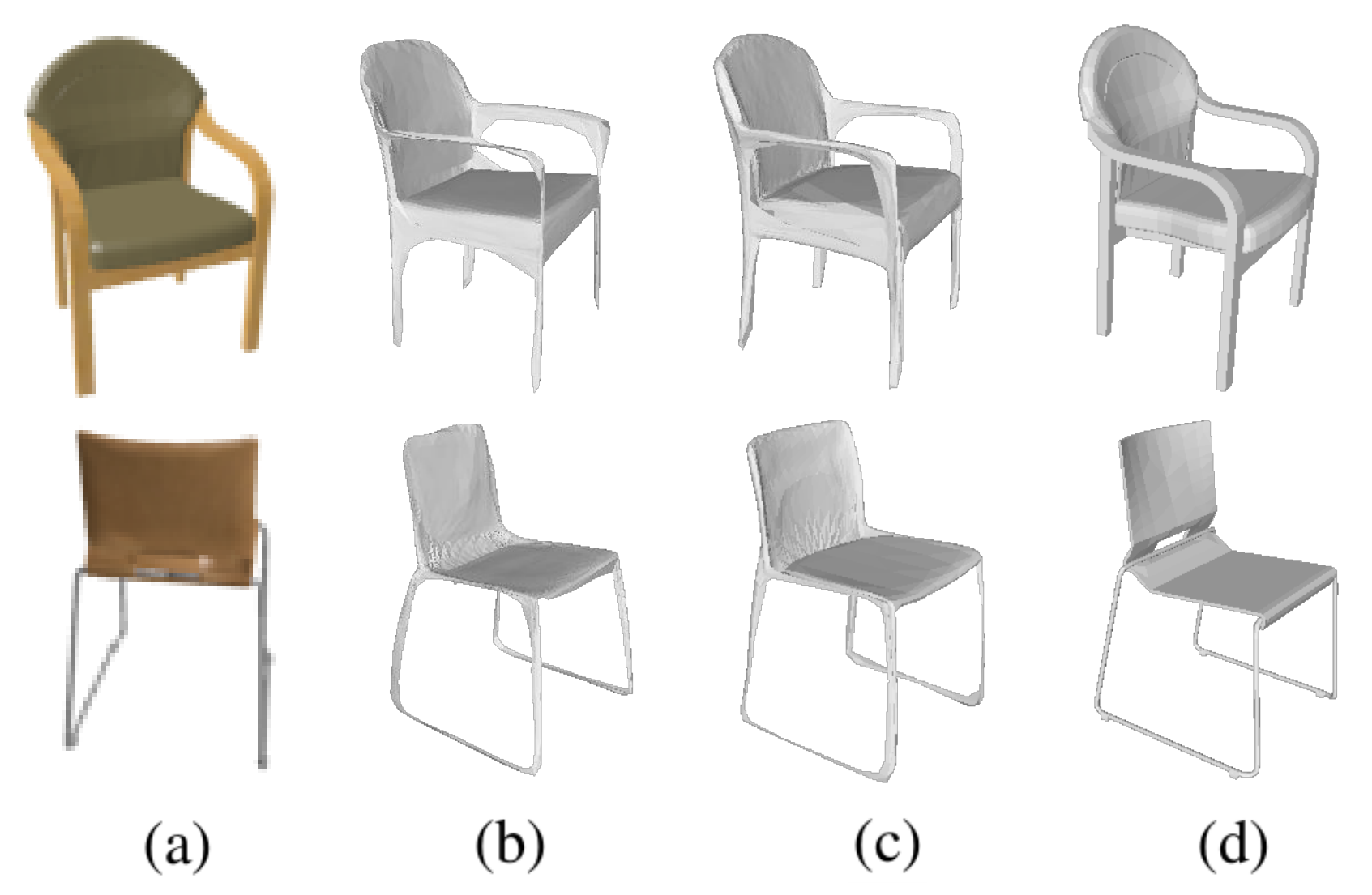}
    \caption{Qualitative results with different initial meshes. (a): Input images; (b): Unit square;
    (c): Sphere; (d): Ground truth.}
    \label{fig:squ_sphe}
\end{figure}

\paragraph{Robustness to initial meshes.}
We first test our approach with different initial meshes (e.g. sphere and unit square).
As seen from the visualization results in Figure~\ref{fig:squ_sphe}, our method achieves similar performance with the two different initial meshes, demonstrating the robustness of our method.

\begin{figure}
    \centering
    \includegraphics[width=0.9\linewidth]{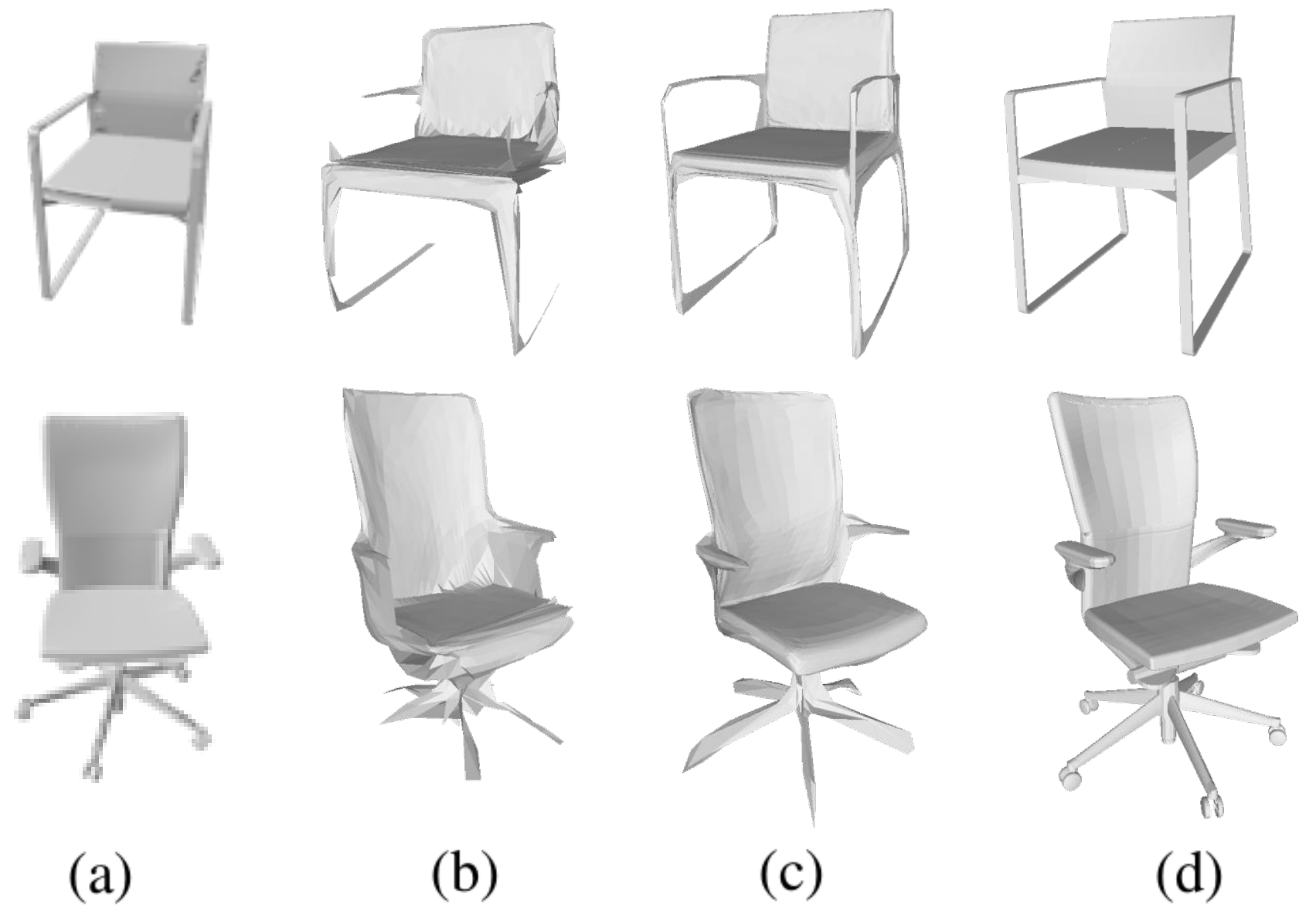}
    \caption{Ablation study on progressive shaping. (a): Input images; (b): Reconstructions w/o progressive shaping; (c): Reconstructions with progressive shaping; (d): Ground truth.}
    \label{fig:pro}
\end{figure}

\paragraph{Progressive shaping.}

Our proposed architecture consists of multiple mesh deformation and topology modification modules that progressively recover the 3D shape.
To validate the effectiveness of such progressive shaping strategy,
we retrain our network with removing the first subnet in the decoder.
Figure~\ref{fig:pro} shows the visualization results. 
Without progressive shaping, the face pruning cannot
be performed in an accurate manner, which could destroy the surface geometry of the generated mesh.


\begin{figure}
    \centering
    \includegraphics[width=1.0\linewidth]{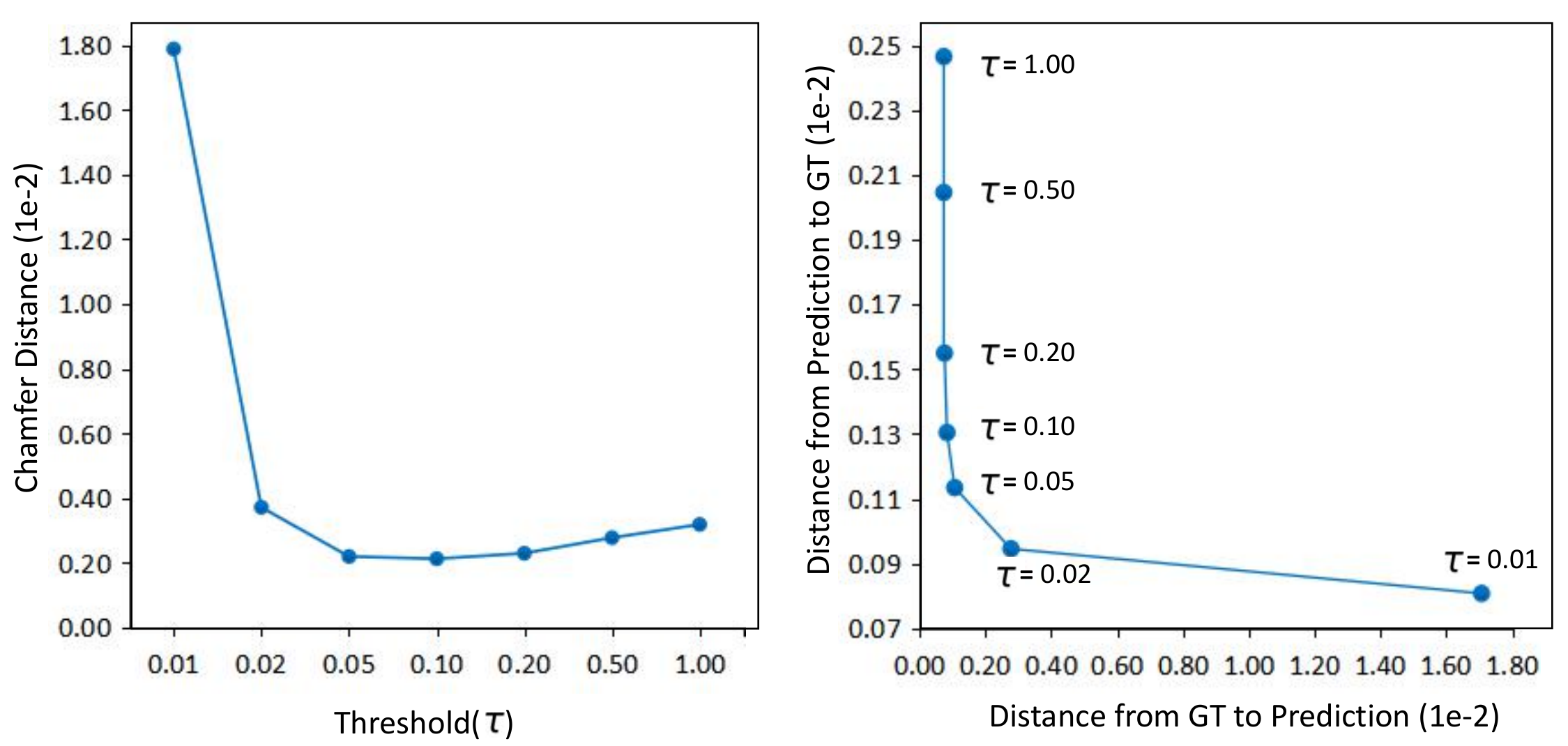}
    \caption{Effect of $\tau$ values on Chamfer distance, distance from GT to prediction, and distance from prediction to GT.}
    \label{fig:tau}
\end{figure}

\begin{figure}
    \centering
    \includegraphics[width=0.95\linewidth]{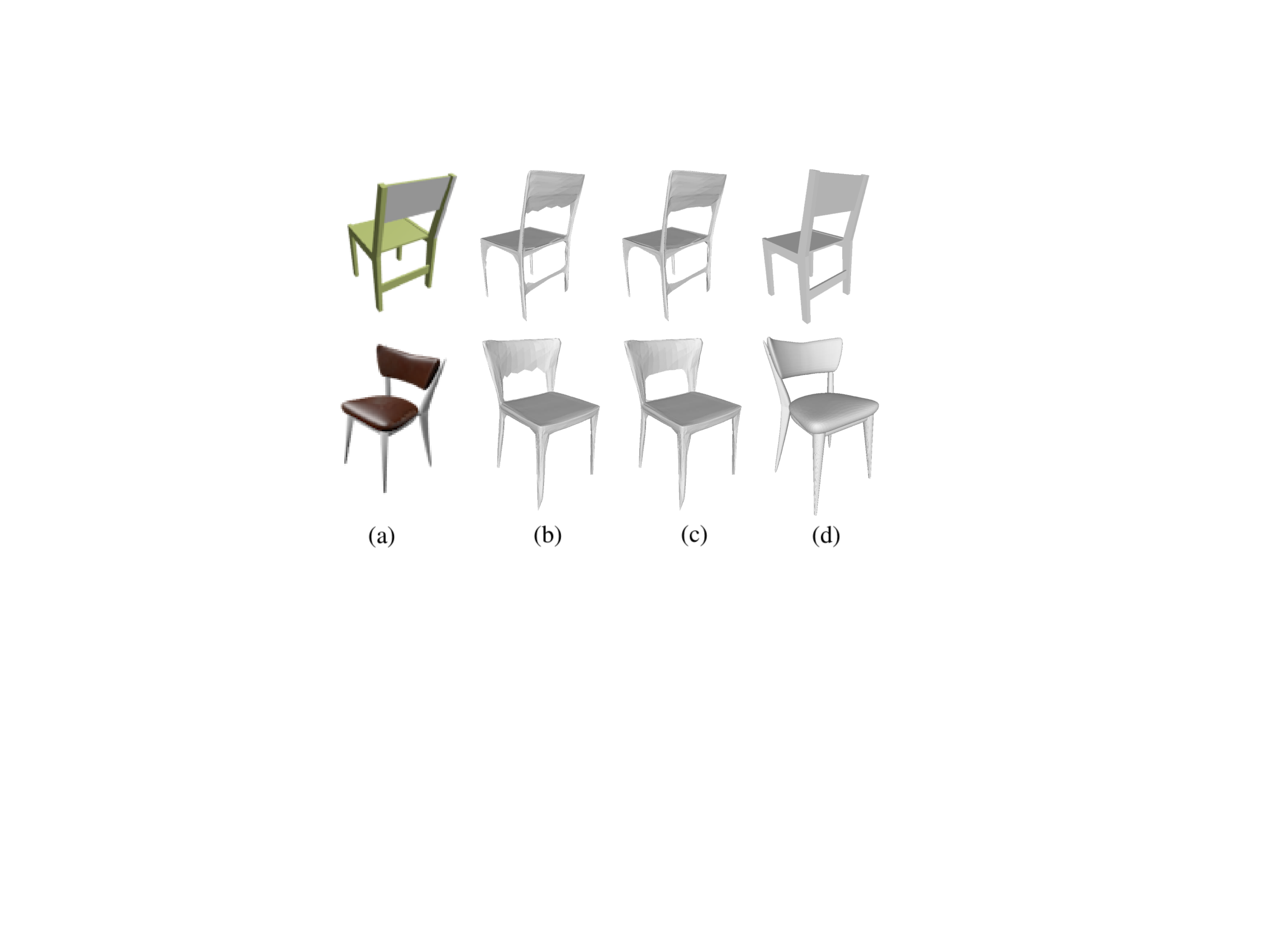}
    \caption{Ablation study on boundary refinement. (a): Input images; (b): Reconstructions w/o boundary refinement; (c): Reconstructions with boundary refinement; (d): Ground truth.}
    \label{fig:reg}
\end{figure}

\paragraph{Face pruning threshold.}

We investigate the effect of the threshold $\tau$ by using $20\%$ of training samples as the validation set, where Chamfer distance (CD) is used as the measure that sums two directional reconstruction errors of $\mathrm{Prediction} \rightarrow \mathrm{GT}$ and $\mathrm{GT} \rightarrow \mathrm{Prediction}$ (cf. Equation (2)). Figure \ref{fig:tau} plots the results, suggesting that $\tau \in [0.05, 0.2]$ strikes a good balance between the two directional distances. We thus set $\tau = 0.1$ in all our experiments.

\paragraph{Boundary refinement.}

To demonstrate the effectiveness of the proposed boundary refinement module,
we show the visualized reconstruction results with and without the boundary refinement in Figure~\ref{fig:reg}.
By using the proposed boundary refinement module, one can achieve significantly cleaner mesh with higher visual quality.



\subsection{Shape Autoencoding}
\begin{figure}
    \centering
    \includegraphics[width=0.93\linewidth]{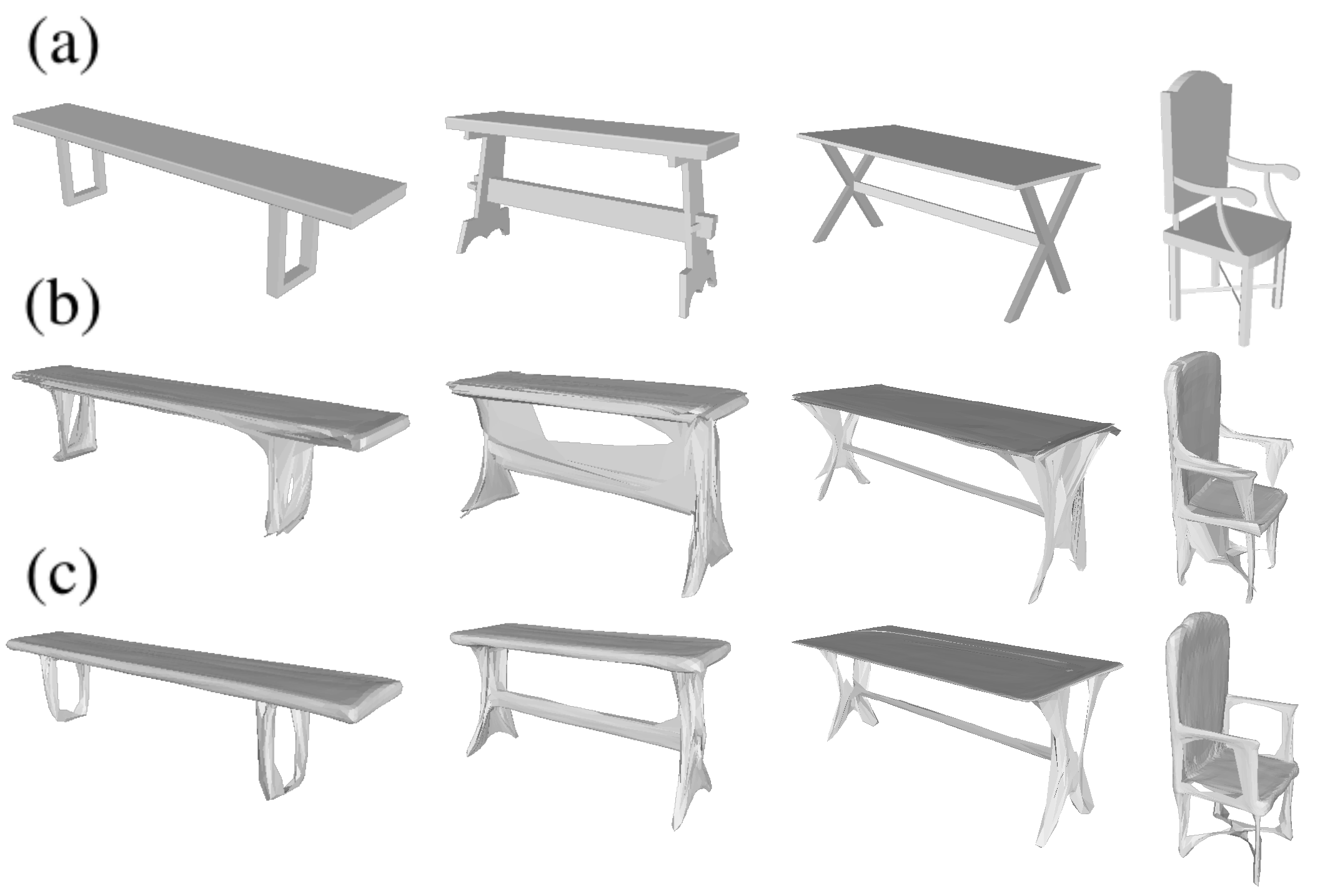}
    \caption{Qualitative results of 3D shape autoencoding. (a): Ground truth Meshes; (b): AtlasNet-25;
    (c): Ours.}
    \label{fig:ae}
\end{figure}

Besides the single-view 3D reconstruction, our framework can also be applied for 3D shape autoencoding.
In this section, we demonstrate the capability of our approach to reconstruct meshes from the input point clouds. 
Toward this goal, we randomly select $2,500$ points from the ground-truth
point cloud and employ the PointNet \cite{qi2017pointnet}  to extract the corresponding latent features. 
Again, we compare both the quantitative and qualitative results
against the state-of-the-art AtlasNet \cite{groueix2018atlasnet}.
To ensure fair comparisons, we use the same experiment settings in \cite{groueix2018atlasnet}.
The results are shown in Table \ref{tab:ae} and Figure \ref{fig:ae}.
As shown in the results, our approach achieves superior performance both qualitatively and quantitatively.

\setlength{\tabcolsep}{10pt}
\begin{table}
    \centering
    \begin{tabular}{c|c|c}
        \hline
         Method & CD    & EMD  \\
         \hline
         AtlasNet-25 (AE)  & 0.765 & 8.467 \\
         Ours (AE)  & \textbf{0.655} & \textbf{6.754} \\
         \hline
    \end{tabular}
    \vspace{0.1cm}
    \caption{Quantitative results of 3D shape autoencoding. The results take the means on the
     five shape categories used in the single-view reconstruction.}
    \vspace{-0.1cm}
    \label{tab:ae}
\end{table}

\section{Conclusion}

We have proposed an end-to-end learning framework that is capable of reconstructing meshes of various topologies from single-view images. 
The overall framework includes multiple mesh deformation and topology modification modules that progressively recover the 3D shape, and a boundary refinement module that refines the boundary conditions.
Extensive experiments show that
our method significantly outperforms the existing methods, both quantitatively and qualitatively.
One limitation of our method is the inherent drawback of producing non-closed meshes. 
But it can be resolved by a post-processing procedure that reconstructs closed surfaces from densely sampled point clouds.
Future research directions include designing a differentiable mesh stitching operation to stitch
the open boundaries introduced by the face pruning operations. 

\section{Acknowledge}
This work is supported in part by the National Natural
Science Foundation of China (Grant No.: 61771201),
the Program for Guangdong Introducing Innovative and
Enterpreneurial Teams (Grant No.: 2017ZT07X183),
the Pearl River Talent Recruitment Program Innovative
and Entrepreneurial Teams in 2017 (Grant No.:
2017ZT07X152),  the Shenzhen Fundamental Research
Fund (Grants No.: KQTD2015033114415450 and
ZDSYS201707251409055), and Department of Science and Technology of Guangdong Province Fund (2018B030338001).

{\small
\bibliographystyle{ieee_fullname}
\bibliography{egbib}
}

\end{document}